\documentclass[10pt,twocolumn,letterpaper]{article}

\usepackage{iccv}
\usepackage{times}
\usepackage{epsfig}
\usepackage{graphicx}
\usepackage{amsmath}
\usepackage{amssymb}
\usepackage{amssymb}
\usepackage{pifont}
\usepackage[font=small,labelfont=bf,tableposition=top]{caption} 
\usepackage{microtype}
\usepackage{xcolor, colortbl}
\usepackage{comment}
\usepackage{float}
\usepackage{siunitx}
\usepackage{amsmath}
\usepackage{enumerate}
\usepackage{xspace}
\usepackage{multirow}
\usepackage{booktabs}
\usepackage[breaklinks=true,bookmarks=false]{hyperref} 
\hypersetup{
    linkcolor=black,
    urlcolor=red,
}

\definecolor{FloorColor}{RGB}{189, 198, 255}

\newcommand{\systemname}{{\sf Neighbor-Vote}\xspace}
\newcommand{\tabincell}[2]{\begin{tabular}{@{}#1@{}}#2\end{tabular}}

\newcommand\blfootnote[1]{%
	\begingroup
	\renewcommand\thefootnote{}\footnote{#1}%
	\addtocounter{footnote}{-1}%
	\endgroup
}

\iccvfinalcopy 

\ificcvfinal\fi
\begin{document}

\title{\systemname: Improving Monocular 3D Object Detection through Neighbor Distance Voting}

\author{
	Xiaomeng Chu $^{1}$ \hspace{0.25cm} Jiajun Deng $^{1}$ \hspace{0.25cm} Yao Li $^{1}$ \hspace{0.25cm}  Zhenxun Yuan $^{2}$ \\ Yanyong Zhang $^{1,*}$ \hspace{0.25cm} Jianmin Ji $^{1}$ \hspace{0.25cm} Yu Zhang $^{1}$
	\vspace{0.1cm}
	\\
	$^{1}$ University of Science and Technology of China \hspace{0.3cm}
	$^{2}$ The University of Sydney
	\vspace{-0.1cm}
	\\
}

\twocolumn[{
	\renewcommand\twocolumn[1][]{#1}%
	\maketitle
}]

\ificcvfinal\thispagestyle{empty}\fi

\begin{abstract}
As cameras are increasingly deployed in new application domains such as autonomous driving, performing 3D object detection on monocular images becomes an important task for visual scene understanding. Recent advances on monocular 3D object detection mainly rely on the ``pseudo-LiDAR'' generation, which performs monocular depth estimation and lifts the 2D pixels to pseudo 3D points.
However, depth estimation from monocular images, due to its poor accuracy, leads to inevitable position shift of pseudo-LiDAR points within the object. Therefore, the predicted bounding boxes may suffer from inaccurate location and deformed shape.
In this paper, we present a novel neighbor-voting method that incorporates neighbor predictions to ameliorate object detection from severely deformed pseudo-LiDAR point clouds. Specifically, each feature point around the object forms their own predictions, and then  the  ``consensus'' is achieved through voting. In this way, we can effectively combine the neighbors' predictions with local prediction and achieve more accurate 3D detection. To further enlarge the difference between the foreground region of interest (ROI) pseudo-LiDAR points and the background points, we also encode the ROI prediction scores of 2D foreground pixels into the corresponding pseudo-LiDAR points.
We conduct extensive experiments on the KITTI benchmark to validate the merits of our proposed method. Our results on the bird’s eye view detection outperform the state-of-the-art performance by a large margin, especially for the ``hard" level detection. 
\end{abstract}

\blfootnote{* Corresponding Author.}

\section{Introduction}
3D object detection is one of the most essential tasks for applications that rely on understanding the context in the 3D physical world, such as  autonomous driving. Nowadays, many 3D detection algorithms have been proposed based on LiDAR point cloud~\cite{voxelnet,pointpillars,Second,pointNet++,yang2018pixor}. Although these approaches achieve impressive performance, however, LiDARs are still too expensive to be equipped on every single vehicle. As a result, inexpensive alternatives are more preferred -- especially cameras, due to their low price and high frame rate.
\begin{figure}[t]
\centering
\begin{tabular}{cc}

\includegraphics[width=0.48\columnwidth]{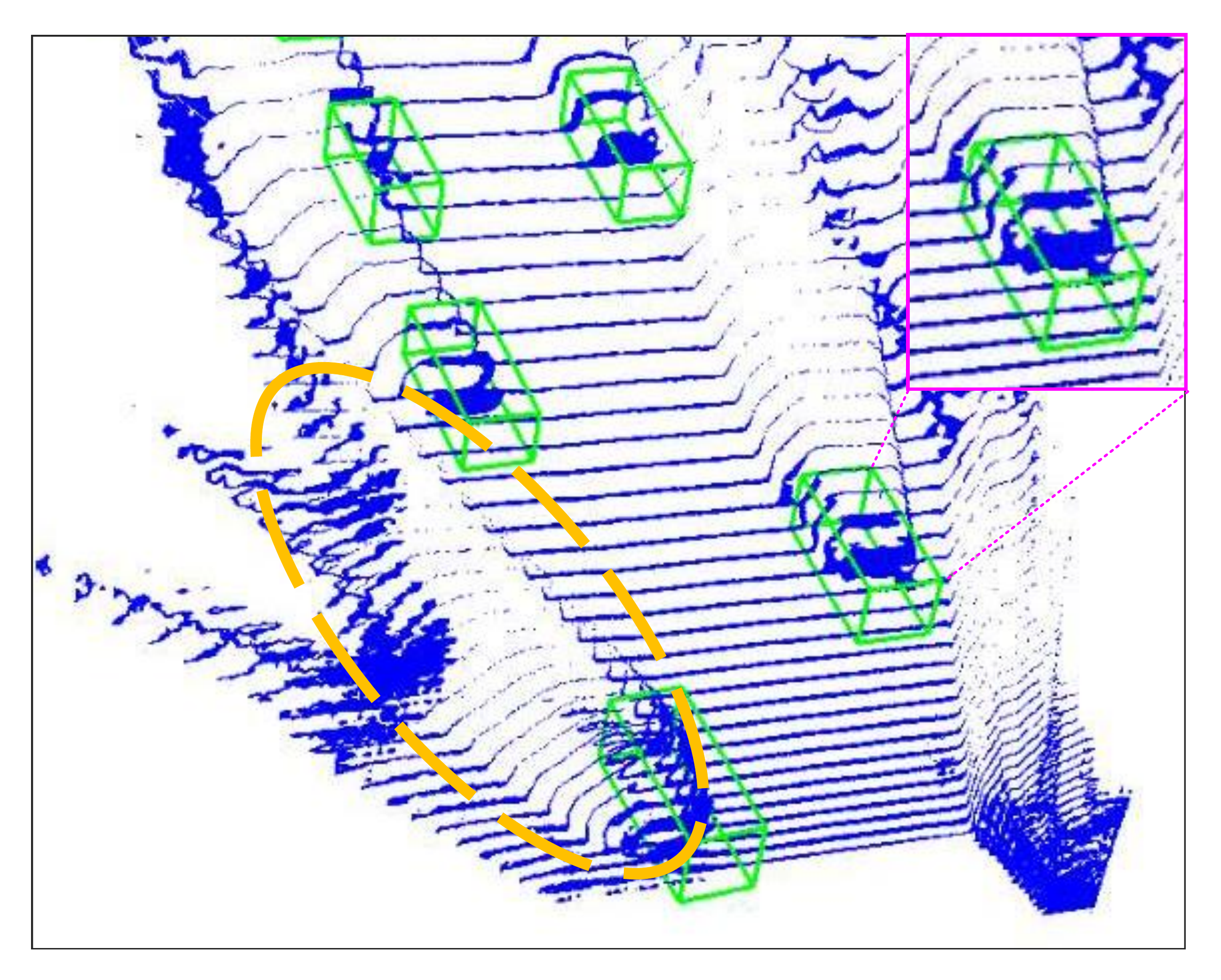}&
\includegraphics[width=0.48\columnwidth]{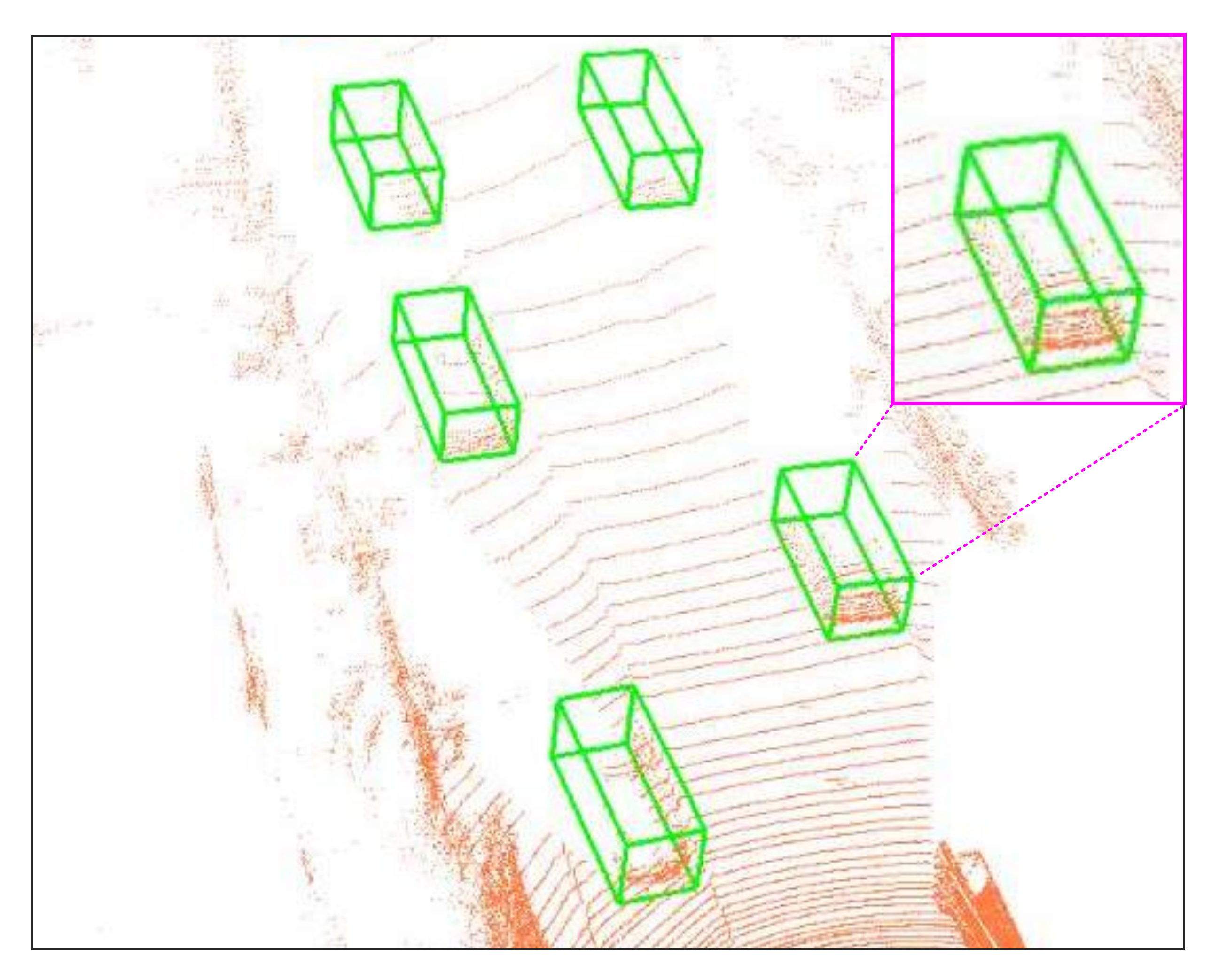} \\ 
(a) pseudo-LiDAR points & (b) LiDAR points \\
\includegraphics[width=0.48\columnwidth]{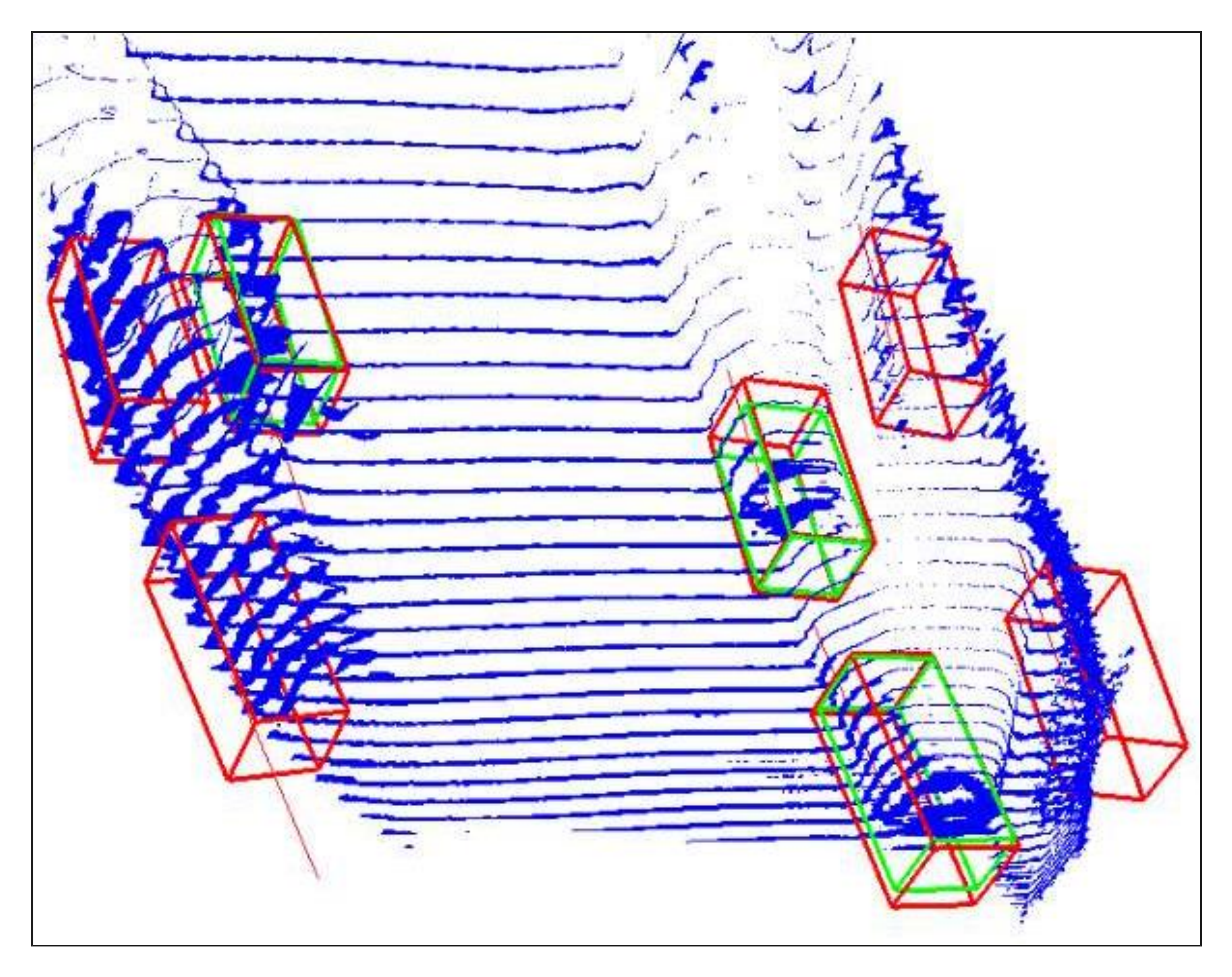} & 
\includegraphics[width=0.48\columnwidth]{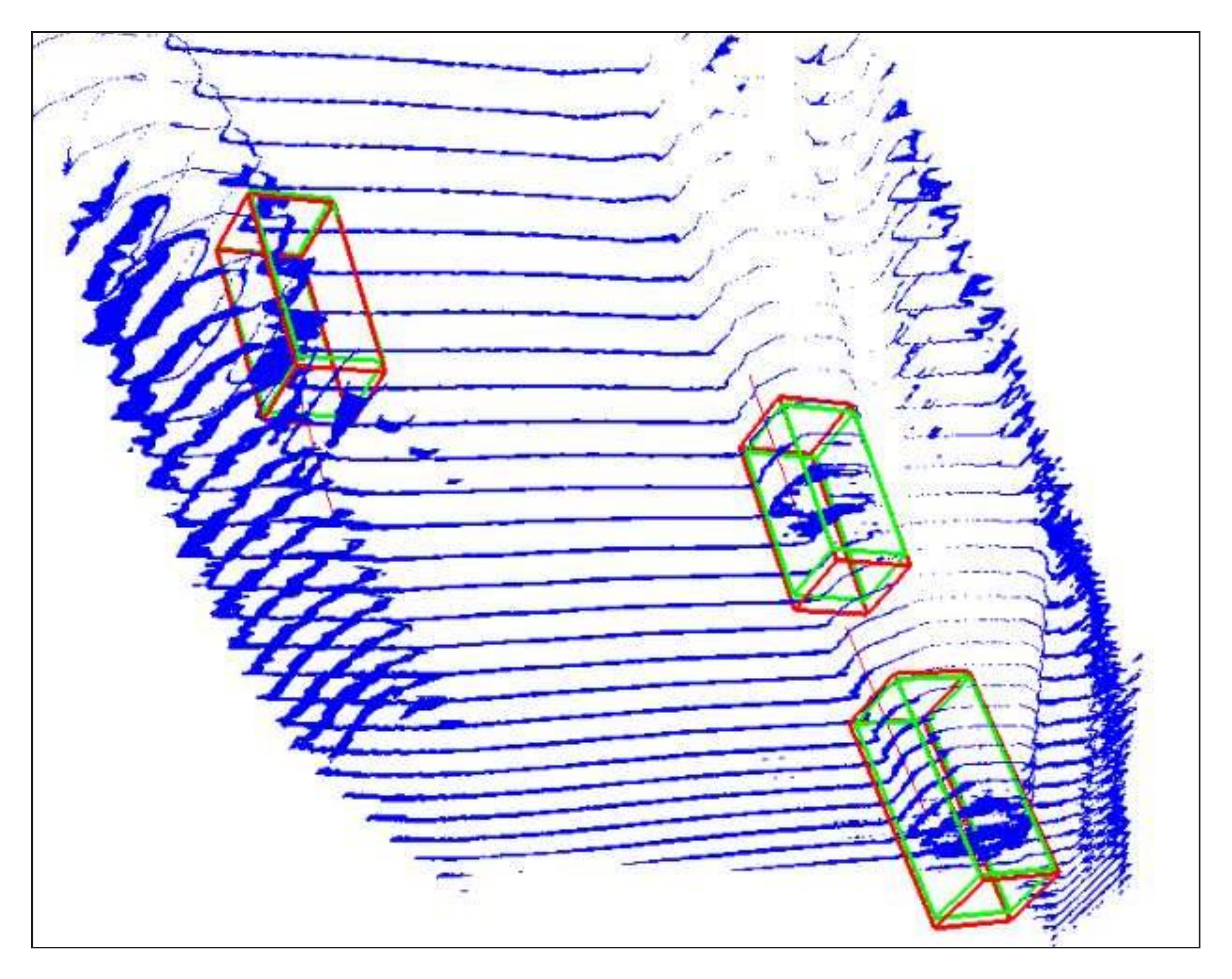} \\ 
(c) baseline box predictions & (d) our predictions \\
\end{tabular}
\caption{\label{fig:intro} (a) and (b) visualize the pseudo-LiDAR and LiDAR point clouds, respectively, in which the details are magnified for better clarity. The pseudo LiDAR point cloud has a long tail marked by the \textcolor{orange}{orange} eclipse. (c) and (d) show the predicted boxes (in \textcolor{red}{red}) in the pseudo-LiDAR point cloud through the baseline network and our work, respectively, in which the ground truth is shown in \textcolor{green}{green}. It shows that our network can effectively eliminate false positive cases generated in the \emph{baseline} network.}  
\end{figure}

On the other hand, conducting 3D detection over RGB images, especially monocular images, remains a daunting task due to the unavailability of depth information~\cite{mousavian20173d,li2020rtm3d,liu2020smoke}. To address this challenge, several methods~\cite{Pseudo-Lidar,weng2019monocular} have proposed to first estimate the depth information from monocular images and then lifts the 2D pixels to pseudo 3D points. Subsequent 3D object detectors can then be applied on the pseudo-LiDAR point clouds.

Compared to LiDAR point clouds, the pseudo-LiDAR obtained as above has unique characteristics that may lead to an array of problems. Below, we point out a few such problems. Firstly, due to the inevitable inaccuracy in depth prediction from monocular images, pseudo-LiDAR often exhibits position shift and shape deformation, which could severely undermine the 3d bounding box regression accuracy~\cite{weng2019monocular,Pseudo-Lidar}. For example, as shown in the orange eclipse of Figure~\ref{fig:intro}(a), the pseudo-LiDAR point cloud often has a $long$ tail which is formed by the depth artifacts at the edge of objects due to the ambiguity of depth estimation.  Secondly, the depth estimation of far objects is less accurate than that of near ones, leading to significantly more distortion
when the object is farther away.

The above observations cannot only explain why pseudo-LiDAR points result in much lower 3D detection accuracy compared to LiDAR points, but also explain why certain false positive cases arise.  For example, objects in the background (\emph{e.g.}, a roadside obstacle)  may likely be regarded as objects of interest (cars in our case) because their pseudo-LiDAR point clouds  are both deformed. 
Considering these false positives that rise from the background, as well as the less defined object shapes with pseudo-LiDAR points, we take the viewpoint that \emph{we need to mine the pseudo-LiDAR data more carefully and pay extra attention to the correlation (or redundancy) embedded in pseudo-LiDAR points}. 
Towards this goal, we propose a method called \textbf{\emph {neighbor-voting}}.
Specifically, we regard each point around the objects in the feature map as a ``voter''. The voter needs to vote for a certain number of nearby objects in its local view. Through this voting process, false positive objects would be included by far fewer votes than true objects, and can thus be more easily identified.  As shown in  Figure~\ref{fig:intro}(d), our method can effectively identify the false positive cases.

As such, the neighbor-voting results can help better predict the object locations. We use them as a new input to generate neighbor-voting classification and then combine the confidence score with the local classification prediction using adaptive weights, which lead to significantly improved detection. The extensive experiments on the KITTI Dataset verify the merits of our approach. Remarkably, our results on both the bird's eye view object detection benchmarks are significantly better than existing monocular methods.

In summary, we make three-fold contributions:
\vspace{-2pt}
\begin{itemize}
    \item We have designed an efficient 3D detection network for monocular images. The network consists of four main steps: pseudo-LiDAR generation, 2D ROI score association, attention-based feature extraction, and neighbor-assisted prediction.
    \vspace{-8pt}\item We have designed a neighbor-voting method that can effectively remove false positive cases in predictions from pseudo-LiDAR point clouds. We can combine neighbor predictions with local predictions in an adaptive manner, which can greatly improve the box prediction accuracy.
    \item Results show our method yields the best performance on KITTI BEV benchmarks, outperforming state-of-the-art methods by noticeable margins.
\end{itemize}

\section{Related Work}
\paragraph{\textbf{LiDAR-based 3D Object Detection.}}

As far as LiDAR-based 3D object detection is concerned, there are two mainstream models: point-based models and voxel-based models.
PointRCNN~\cite{pointrcnn} is a point-based detector. It extracts features from raw LiDAR point clouds through PointNet++~\cite{pointNet++} and sends the features to RPNs networks for further detection.
VoxelNet~\cite{voxelnet} demonstrates a powerful voxel feature encoding layer that converts 3D LiDAR point clouds into equally spaced 3D voxels. SECOND~\cite{Second} utilizes the voxel feature extraction layer and introduces a novel sparse convolution layer targeted at LiDAR point clouds. 
PointPillars~\cite{pointpillars} uses pillars rather than voxels to encode point cloud features, and further improves the network efficiency. Voxel R-CNN~\cite{voxel-rcnn} proposes voxel query and voxel RoI pooling, which takes full advantage of voxel features in a two stage approach. PV-RCNN~\cite{pvrcnn} combines the advantages of points and voxels, which devises voxel set abstraction to integrates multi-scale voxel features into sampled keypoints.

\begin{figure*}[t]

    \centering
    \includegraphics[width=.95\textwidth]{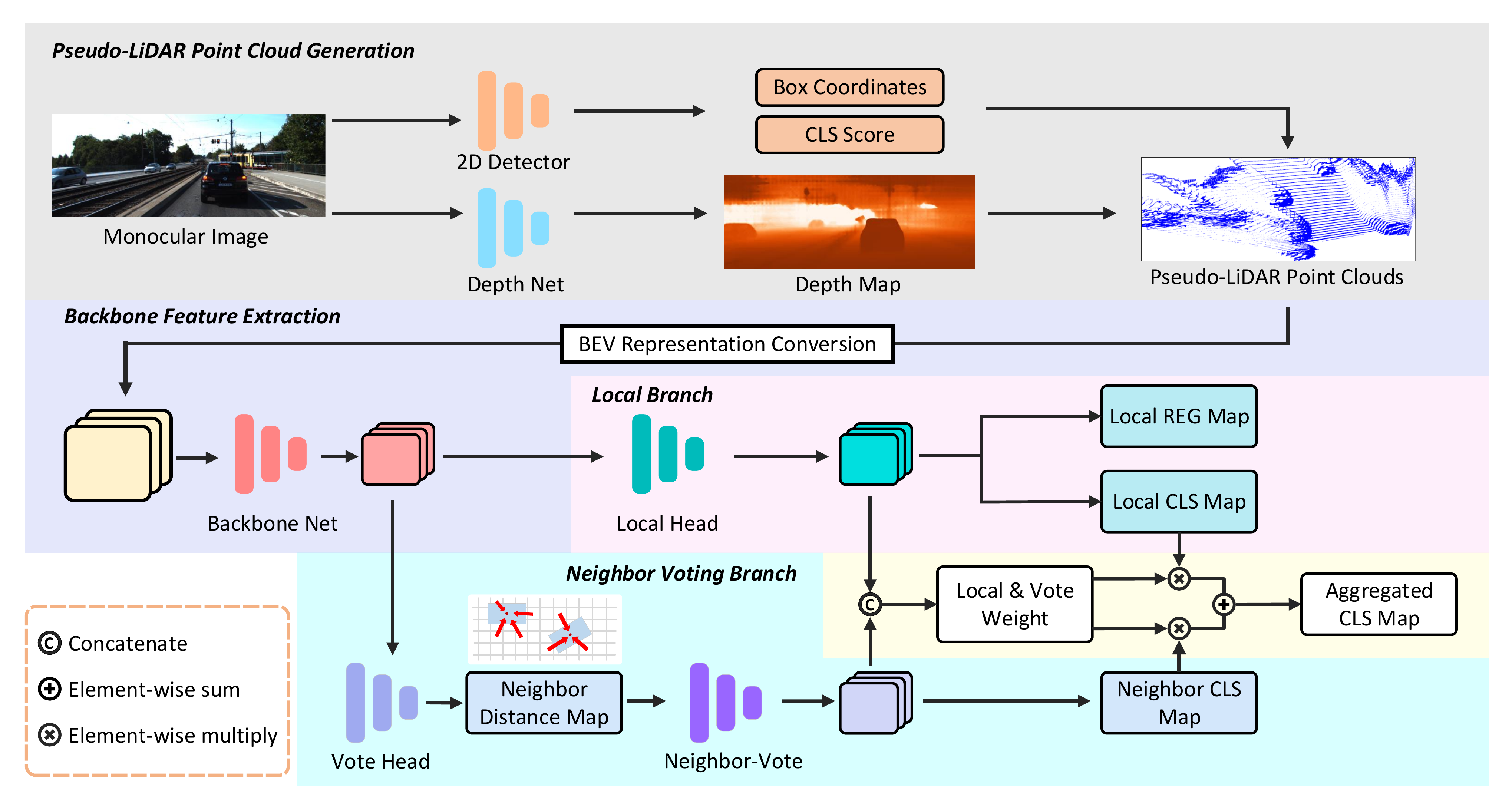}

    \vspace{-6pt}\caption{Overview of the \systemname\ pipeline. The 2D monocular image is first transformed into the 3D representation through a pseudo-LiDAR point clouds generation module. A 2D Backbone with self-attention module is then adopted to extract the features on the bird's eye view. The neighbor-voting process is integrated into the pipeline. Neighbor votes and local classification are combined adaptively through the fusion of the two branches. } 
    \label{fig:pipeline}
\end{figure*}

\paragraph{\textbf{Monocular 3D Object Detection.}}

In recent years, a large number of monocular 3D detection algorithms have emerged. 
Mousavian et al.~\cite{mousavian20173d} utilizes the deep learning network and geometric constraints of 2D Bounding boxes to generate 3D Bounding Boxes and a pose with 6 degrees of freedom.
Li et al.~\cite{li2020rtm3d} regresses 3D boxes through predicting their nine perspective keypoints in image space. There is still a big gap between LiDAR-based and image-only based methods in the 3D detection accuracy. A lot of work introduces other information, such as depth, semantic segmentation, CAD models, \emph{etc.}, which play a role in improving the results.
Ding et al.~\cite{ding2020learning} proposed a new local convolutional network, where the filters are generated from the depth map corresponding to the input image. The filters and their receptive fields are dynamically different in different images and different pixel positions.
Y.Wang et al.~\cite{Pseudo-Lidar} proposed a new data representation ``Pseudo-LiDAR", which is obtained from the combination of image pixel coordinates and corresponding depth. Weng et al.~\cite{weng2019monocular} adopts this data representation and use a 2D-3D bounding box consistency constraint for better predictions.  AM3D~\cite{ma2019accurate} combines 2D detection and monocular depth estimation to process the image to obtain 2D Box and depth map. Then the 3D point clouds are generated to regress 3D Boxes. Pseudo-LiDAR effectively narrows the gap between image-based and LiDAR-based 3d object detection. 

\vspace{-0.45cm}
\paragraph{\textbf{Vote-based 3D object detection.}}
A few 3D object detection methods have been proposed to incorporate voting ideas. VoteNet~\cite{votenet} proposes to endow point cloud deep networks with a voting mechanism similar to the classical Hough voting that is based on point set networks. MLCVNet~\cite{MLCVNet} exploits multi-level contextual information and fuses them to vote for the object center. MLCVNet++~\cite{MLCVNet++} further improves the network by proposing 3DNMS to remove redundant detections during post-processing. Yan et al.~\cite{fuse_vote} combine texture information from RGB data and geometric information from point cloud data, and use the deep Hough voting algorithm to suggest objects. 
The motivation for voting in these methods is ``3D object centroid can be far from any surface point captured by depth sensors". Most of them use LiDAR points as ‘seeds’. In contrast, we use group predictions to remove false positive cases, and our voters are extracted feature points.

\section{The \systemname Design}

\subsection{Overview}
In this paper, we propose \systemname, a pseudo-LiDAR based framework aiming at improving monocular 3D object detection with additional object location estimations from neighbors. The main innovation of this work lies in the design of our neighbor-voting module, which takes into consideration the box predictions from neighbors to ameliorate object detection from severely deformed pseudo-LiDAR. As illustrated in Figure~\ref{fig:pipeline}, our proposed \systemname adopts the one-stage paradigm, and consists of the following four main steps:

\vspace{-2pt}

\begin{enumerate}[(1)]
    \item Pseudo-LiDAR Generation. This step generates the pseudo-LiDAR point clouds from monocular images. We first obtain the depth map through a depth estimation model. We then combine the map with pixel coordinates to create pseudo-LiDAR point clouds as described in~\cite{Pseudo-Lidar}.

    \item 2D ROI Score Association. This step encodes the scores of predicted 2D boxes into the corresponding pseudo-LiDAR points, which are projected from the depth map in the boxes.

    \item Attention-Based Feature Extraction. This step combines the short-range and long-range information. Here, we employ a self-attention module~\cite{Attention,non-local} with a 2D convolutional backbone derived from PointPillars~\cite{pointpillars} on the BEV, extracting both non-local features and multi-level local features.

    \item Neighbor-Assisted Box Prediction. This step incorporates additional object location information  from neighbors to assist with box localization from translated and deformed pseudo-LiDAR. The network is split into two master branches: neighbor-voting branch and local branch. The outputs from these two branches are later combined by applying adaptive weights.

\end{enumerate}
We show the overall pipeline in Figure~\ref{fig:pipeline}, and discuss the four steps one by one in the following subsections. 
 
\subsection{Pseudo-LiDAR Generation from Estimated Depth Map}
\label{sec:pseudo-gene}
\systemname\ can employ a variety of depth estimation models. In this paper, we use the DORN model (originally proposed in~\cite{fu2018deep}) in the same way as in~\cite{Pseudo-Lidar,weng2019monocular} for its high accuracy and low root-mean-square error (RMSE). After the depth estimation, we then combine an image pixel's coordinate $(u, v)$ with its depth to generate a pseudo-LiDAR point $(x, y, z)$\footnote{We use camera coordinate system in this paper, where $X$ axis is right, $Y$ axis is down, $Z$ axis is forward when you face to the scenes.}. The transformation can be formulated as follows:
\begin{equation}
\centering
    x = \frac{(u - c_x)z}{f_x}, \quad
    y = \frac{(v - c_y)z}{f_y}, \quad
    z = Z(u, v),
\end{equation}
where $(c_x, c_y)$ is the camera center in pixel location, and $z$ is the pixel's estimated depth. After the above 2D-to-3D data representation transformation, LiDAR-based detection algorithms can be applied subsequently to process the pseudo-LiDAR points.

\subsection{Foreground Pseudo-LiDAR Point Likelihood Association}
The depth estimation accuracy for faraway objects is much lower than that for near objects, resulting in worse position shift in the pseudo-LiDAR points at greater distances. To compensate the inaccurate depth estimation, in this step we endeavor to enlarge the difference between the foreground region of interest (ROI) and the background, especially for faraway objects. For this purpose, we propose to associate each foreground 2D pixel's ROI score to the corresponding pseudo-LiDAR point, using the score to indicate the likelihood of being a foreground point.

We find that in 2D images,  a faraway object, though small and low-resolution, usually still retains a certain level of semantic information. In fact,  as far as the ``car" category in KITTI~\cite{KITTI} Dataset is concerned, the average precision (AP) has reached above $75\%$ on the ``hard" level objects with the IoU threshold of 0.7 for many 2D detectors, \emph{e.g.} FCOS~\cite{FCOS}, CenterNet~\cite{Centernet}, Cascad R-CNN~\cite{cascade}. 
With this observation, we propose to have a 2D detector extract the ROI areas and  associate the prediction score with the corresponding pseudo-LiDAR points. 

Our 2D detector is derived from FCOS~\cite{FCOS}. We use each pixel's score of the predicted bounding box, and project it into the 3D space as explained in Sec.~\ref{sec:pseudo-gene}. We then have this score encoded as the fourth channel of the pseudo-LiDAR points as shown in Eqn.~\eqref{equ:confidence}:
\begin{equation}
\centering
p_c = \begin{bmatrix} x \\ y \\z \\\sigma \end{bmatrix}.
\label{equ:confidence}
\end{equation}

\subsection{Feature Extraction with Self-Attention}
\begin{figure}[t]
\centering
\begin{tabular}{ccc}

\includegraphics[width=0.28\columnwidth]{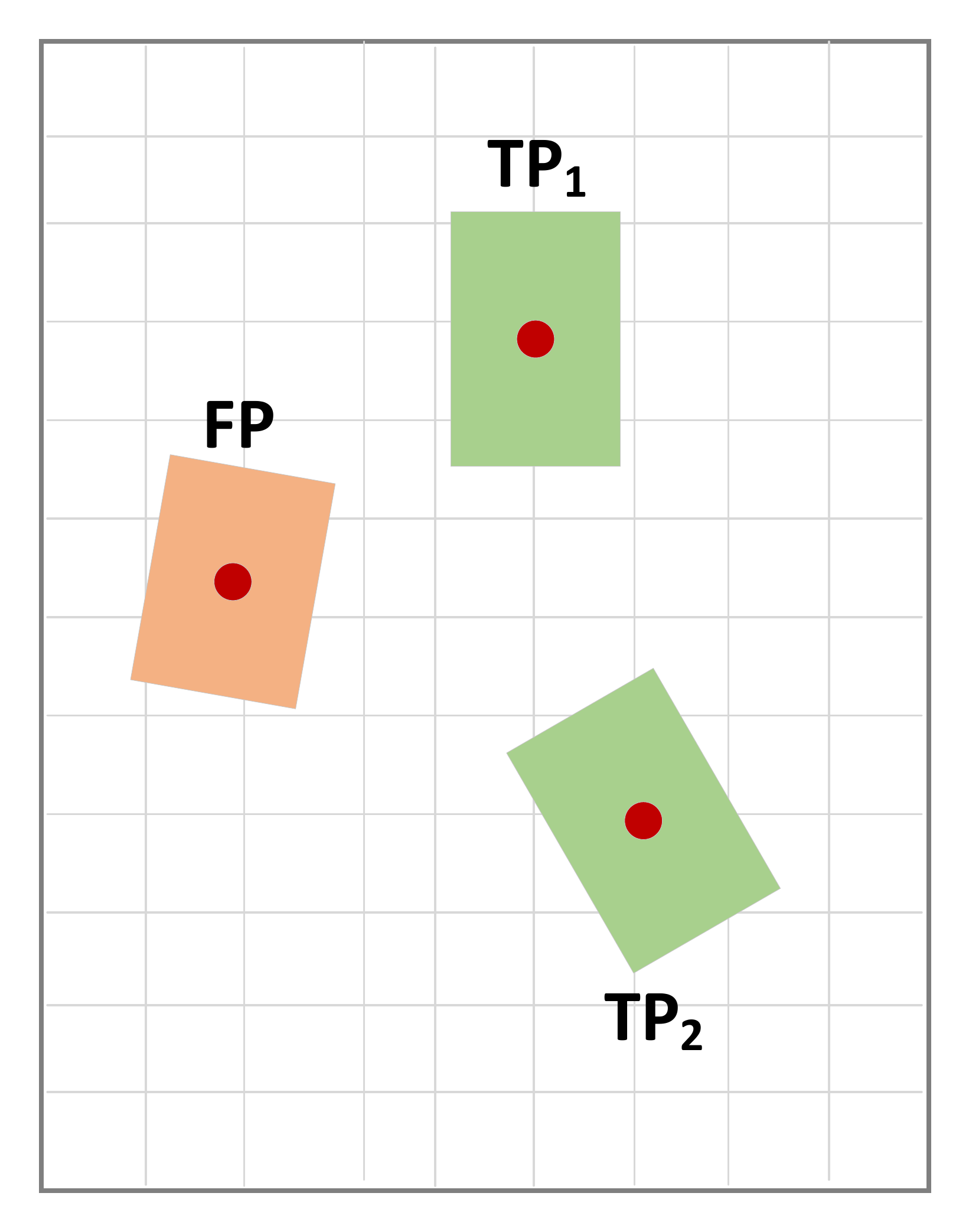}&
\includegraphics[width=0.28\columnwidth]{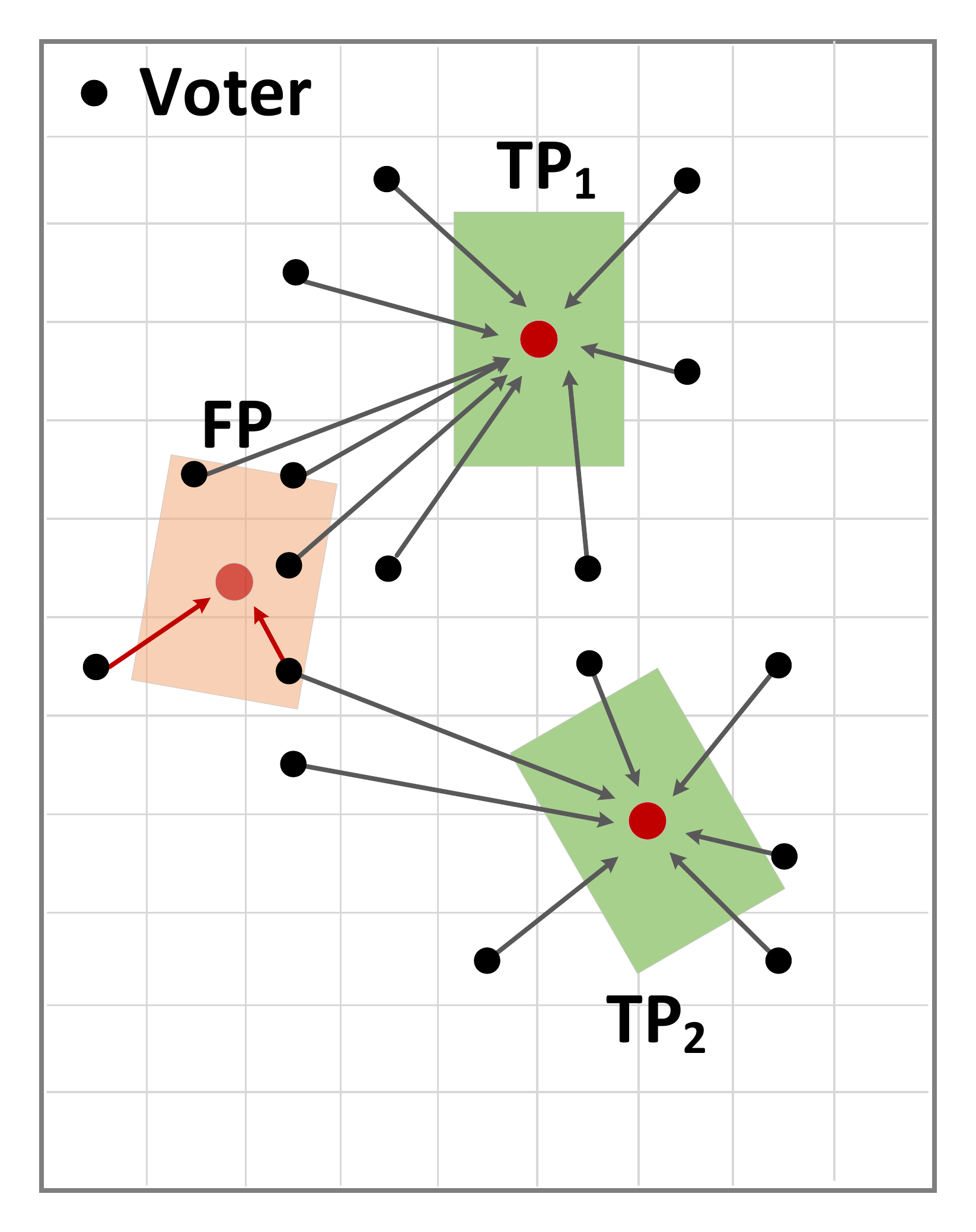}& 
\includegraphics[width=0.28\columnwidth]{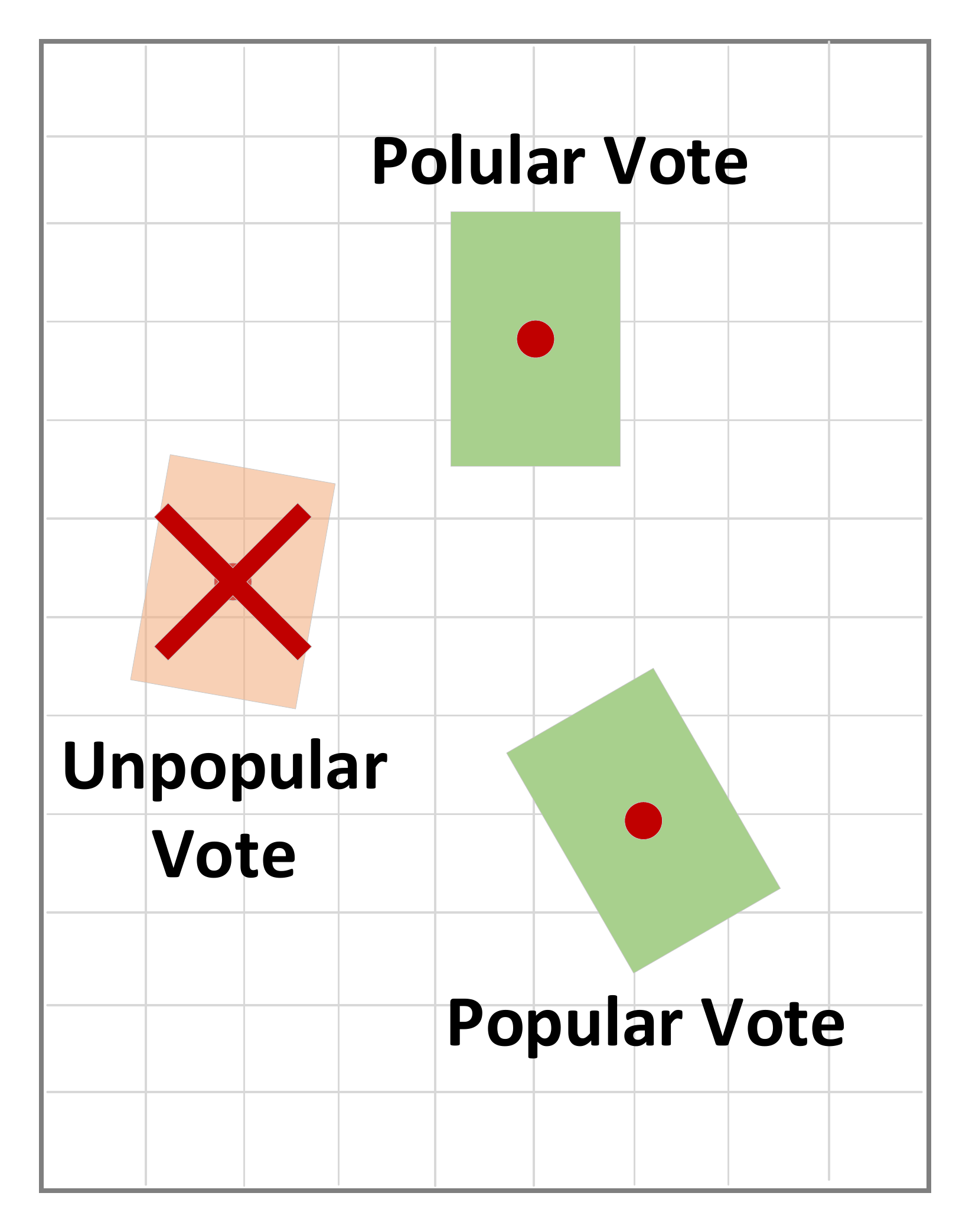} \\
(a) & (b) & (c)
\end{tabular}
\caption{\label{fig:reason} Illustration of the voting process. (a) Supposing we have two true positive predictions $TP_1$, $TP_2$ (in \textcolor[RGB]{168,208,141}{green}) and one false positive prediction $FP$ (in \textcolor[RGB]{244,177,131}{orange}). (b) The feature points (voters) around objects vote for the cars they think are the closest. (c) True positives $TP_1$ and $TP_2$ receive more votes than $FP$, $FP$ then be removed.}
\end{figure}
\paragraph{\textbf{2D Feature Extraction.}}

As with many 3D detectors~\cite{voxelnet,Second,pointpillars,yang2018pixor}, we first voxelize the pseudo points into grids. The range of the point clouds is limited to  $L\times W\times H$, and is divided into $l\times w\times h$ grids as the input of the network. The size of each voxel cell is $\frac{L}{l}\times\frac{W}{w}\times\frac{H}{h}$, where $h$ is regarded as the channel of the input. We adopt the backbone and detection head of PointPillars~\cite{pointpillars} with $h=1$. Because the density of the pseudo-LiDAR points is much higher than that of the LiDAR points, we downsample the points 6 times and set the maximum number of points per voxel to 4 times that for LiDAR points.

\paragraph{\textbf{Spatial Context Retrieval.}}

Due to the serious displacement and deformation of the pseudo-LiDAR point clouds, the spatial context of each feature point that relies upon the long-range information is needed to better identify the position and shape of the object.  Stacks of convolutional operations with fixed receptive fields at each position cannot extract sufficient long-range features effectively. We thus apply the self-attention mechanism in our feature extraction and neighbor-vote module.

We elaborate on our self-attention module below. The feature map input is transformed and flattened into three vectors $Q\in \mathbb{R}^{c_Q\times N}$, $K\in \mathbb{R}^{c_K\times N}$ and $V\in \mathbb{R}^{c_V\times N}$, where $c_Q$, $c_K$ and $c_V$ represent the channels of three vectors respectively, and $N$ is the number of spatial positions. The spatial context of a feature point is obtained by the weighted sum of itself and all other feature points as follows:

\begin{equation}
    o_i = \sum_{j=1}^N w_{i,j} \psi(x_j),
\vspace{-0.1cm}
\end{equation}
where $o_i$ is the output context at position $i$ and $w_{i,j}$ is the normalized attention weight. $\psi$ indicates a FC layer that transforms $x$ (the flattened feature map) into value space. The weight map $W\in \mathbb{R}^{N\times N}$ is computed as:
\begin{equation}
    W = softmax(\frac{Q^T \times K}{\sqrt{c_K}}).
\vspace{-0.1cm}
\end{equation}

\subsection{Box Prediction with Neighbor-Voted Object Locations}
\paragraph{\textbf{Neighbor Voting.}} 
As mentioned before, the pseudo-LiDAR points are not as accurate as actual LiDAR points in terms of depicting the object position and shape. In order to address this challenge, we propose to take advantage of the feature points near the object (which we refer to as ``neighbors'') and have them assist in judging the object location. Specifically, we leverage the individual perspective of each neighbor point and try to form a ``consensus'' through a voting mechanism. Let us consider a BEV feature map $(L_\lambda\times W_\lambda)$, where $L_\lambda$ and $W_\lambda$ represent the size of feature map in $x$ and $z$ directions respectively, and $\lambda$ represents the down-sampling rate. Feature points that are close to a predicted object are regarded as voting neighbors, or `voters'.
Each voter casts two votes. Namely, they vote for the two closest objects, one in \textbf{front} and one in \textbf{back} (relative positioning in the $z$ direction), denoted by $(\theta_{f}, dz_{f}, \theta_{b}, dz_{b})$, where $\theta$ is the the included angle between the object and the positive $X$-axis from the perspective of voters, and $dz$ is the distance from the object in $Z$ directions.
From the angle and distance information, we can get the exact positions of the two objects for each voter.

Please note that the voter compares the predicted object center location $(x_c, z_c)$ with its own location $(x_v, z_v)$ in the $z$ direction to determine the relative positioning (front or back). When choosing which object is the closest to the voter, it uses the 2D real euclidean distance:
\begin{equation}
\begin{aligned}
	C_{f} = \mathop{\arg\min}_{c} \sqrt{(x_c-x_v)^2 + (z_c-z_v)^2}&,\\
	 c \in P\ and\ z_c <=z_v&, \\
	C_{b} = \mathop{\arg\min}_{c} \sqrt{(x_c-x_v)^2 + (z_c-z_v)^2}&,\\
	c \in P\ and\ z_c > z_v&,    
\end{aligned}
\label{equ:choosecar}
\end{equation}
where $P$ is the list of predicted objects. $C_f$ and $C_b$ are the chosen objects in front and back. Here, we first have all the features points participate in voting, and then filter out those whose votes are beyond a certain distance. In this way, all the voting neighbors are indeed near predicted objects.

With a simple vote head, we can obtain the neighbor distance map from the votes.
The 6-channel neighbor distance map contains the individual prediction of each feature point, namely, $$\left\{sin(\theta_f), cos(\theta_f), dz_{f}, sin(\theta_b), cos(\theta_b), dz_{b}\right\}.$$ Then, the map can be used to achieve the detection ``consensus'' in the neighbor-voting process.

Figure~\ref{fig:reason} shows a neighbor-voting example. There are three predictions, with two true positive predictions ($TP_1$ and $TP_2$ shown in green) and one false positive prediction ($FP$ shown in orange). The rationale here is that most voters around $TP_1$ and $TP_2$ vote them as the closest cars, while most feature points around $FP$ do not vote for $FP$. Based upon this rationale, the classification score for $FP$ will be much lower and the false positive case can be eliminated. We believe in this rationale because \emph{most of the prediction errors are not systematic errors, but have some degree of randomness and uncertainty.} As a result, the number of feature points that make the same prediction errors are not as many as those that make the same correct predictions. 
Next, we briefly validate our rationale. We first count the feature points that vote for a FP prediction --  among all the feature points that are within 6 meters from a FP prediction, $39.4\%$ vote for the FP object. Meanwhile, we find that  $72.3\%$ of the feature points that are within 6 meters from a TP object vote for the TP object. This agrees with our rationale that true positive cases receive significantly more ``votes'' than false positive cases.
\label{sec:nv}

\paragraph{\textbf{Combination of Neighbor Voting and Local Classification.}}

We next explain how we fuse these two classification results from local branch and neighbor-voting branch for the final classification result. 
Both branches produce a $k$-channel output, $P_{local}$ and $P_{vote}$, respectively, where $k$ is the number of anchors in each position. We then concatenate the features from neighbor-voting branch and local branch and then apply a softmax function, obtaining a two-channel weight map. The values of these two channels are named $W_{local}$ and $W_{vote}$, which sum up to 1 for each position. The final weighted classification score is computed as follows:
\begin{equation}
P_{fusion} = W_{local}\cdot P_{local} + W_{vote}\cdot P_{vote}. 
\label{exp:p_final}
\end{equation}

\subsection{Training Losses}
The overall loss is composed of two parts: one is for the detection objective, including the classification loss $L_{cls}$ and box regression loss $L_{reg}$, and the other is for neighbor-voting, including $L\_{dist}$ and $L\_{ang}$. The formula is as follows:
\begin{equation}
    Loss = \lambda_{det} L_{det} + \lambda_{nv} L_{nv}
\end{equation}

The value of $L_{cls}$ is the weighted sum of local classification loss $L_{local}$, neighbor-voting classification loss $L_{vote}$, and fusion classification loss $L_{fusion}$. 
To stabilize the training process, we adopt three hyper parameters as their respective coefficients to re-weight these losses:
\begin{equation}
    L_{cls} = \alpha\cdot L_{local} + \beta\cdot L_{vote} + \gamma\cdot L_{fusion},
\end{equation} 
with $\alpha=1$, $\beta=1$, and $\gamma=2$ in the training phase. 

We use focal loss~\cite{focal_loss} on all the classification output and smooth $l_1$ loss on all regression output.

\section{Implementation and Evaluation}
\label{experiments}
In this section, we present our implementation details and evaluation results.

\begin{table*}[tb]

\centering
\small
\caption{Performance Comparison on KITTI \emph{val} set in terms of $AP_{BEV}$ and $AP_{3D}$. ``Extra Info'' means extra supervision in addition to the ground-truth boxes, where ``mask" means the label of segmentation task. \systemname\ outperforms the state-of-the-art monocular networks by noticeable margins. }
\setlength{\tabcolsep}{4.5pt}
\begin{tabular}{l|cc|rrr|rrr|rrr|rrr}
\toprule
    \multirow{2}{*}{Method} &\multicolumn{2}{c|}{Extra Info} &
    \multicolumn{3}{c|}{$\text{AP}_{\text{BEV}}$ (\%),
    \textbf{IoU = 0.5} } & \multicolumn{3}{c|}{$\text{AP}_{\text{BEV}}$ (\%),
    \textbf{IoU = 0.7}} &
    \multicolumn{3}{c|}{$\text{AP}_{\text{3D}}$ (\%),
    \textbf{IoU = 0.5} } & \multicolumn{3}{c}{$\text{AP}_{\text{3D}}$ (\%),
    \textbf{IoU = 0.7}}\\
    &Depth&Mask& Easy & Mod. & Hard & Easy & Mod. & Hard & Easy & Mod. & Hard &Easy & Mod. & Hard \\
    \hline
    Deep3DBox~\cite{mousavian20173d} & & & 30.0 & 23.8& 18.8 & 10.0 & 7.7 & 5.3& 25.2& 18.2& 15.5& 2.5& 2.3 & 2.3\\

    SMOKE~\cite{liu2020smoke}& & & - & -& - & 20.0 & 15.6 & 15.3& -& -& -& 14.8& 12.9& 11.5\\ 
    OAV~\cite{OAV}& & & 51.2 & 38.3 & 34.3 & 20.7 & 16.4 & 14.2& 44.7& 32.8& 28.3& 13.7& 11.5& 10.7\\   
    RTM3D(DLA34)~\cite{li2020rtm3d}& & & 57.5& 44.2& 42.3 & 25.6 & 22.1 & 20.9& 54.4& 41.9& 35.8& 20.8& 16.9& 16.6\\ 
    \hline
    ROI-10D~\cite{manhardt2019roi}&\checkmark& & 46.9 & 34.1& 30.5& 14.5 & 9.9& 8.7& 37.6& 25.1& 21.8& 9.6& 6.6& 6.3\\ 

    MonoGRNet~\cite{qin2019monogrnet}&\checkmark& & 54.2 & 39.7 & 33.1 &25.0 & 19.4 & 16.3& 50.5& 37.0& 30.8& 13.9& 10.2& 7.6\\   
    D4LCN~\cite{ding2020learning}&\checkmark& & - & - & - & 34.8 & 25.8 & 23.5& -& -& -& 27.0& 21.7& 18.2\\ 

    PL-MONO~\cite{Pseudo-Lidar}&\checkmark& & 70.8 & 49.4 & 42.7 & 40.6 & 26.3 & 22.9& 66.3& 42.3& 38.5& 28.2& 18.5& 16.4\\ 

    AM3D~\cite{ma2019accurate}&\checkmark&\checkmark& 72.6 & 51.8 & 44.2 & 43.8 & 28.4 & 23.9& 68.9& 49.2& 42.2& 32.2& 21.1& 17.3\\ 

    Mono3D-PL~\cite{weng2019monocular}&\checkmark&\checkmark& 72.1 & 53.1 & 44.6 & 41.9 & 28.3 & 24.5& 68.4& 48.3& 43.0& 31.5& 21.0& 17.5\\ 
  
    PatchNet~\cite{ma2020rethinking}&\checkmark& & - & - & - & 44.4 & 29.1 & 24.1& -& -& -& \textbf{35.1}& 22.0& 19.6\\ 

    \hline
    \hline

    \textbf{\systemname (ours)}&\checkmark & &\textbf{74.0} & \textbf{55.1} & \textbf{48.1} &
    \textbf{48.2}& \textbf{33.4} & \textbf{28.2}& \textbf{71.6}& \textbf{48.5}& \textbf{45.8} & 33.4 & \textbf{23.4} & \textbf{20.1}\\

\bottomrule
\end{tabular}

\label{table:eval_bev}
\end{table*}

\subsection{Implementation Details}

\paragraph{\textbf{Dataset.}} All our experiments are evaluated on the KITTI object detection benchmark~\cite{KITTI}, which contains 7,481 images for training and 7,518 images for testing. We split the 7,481 samples into 3,712 and 3,769 for training and validation respectively as in~\cite{chen20153d}. We compare our network with existing methods on the \emph{test} set by training our model on both the \emph{train} and \emph{val} sets.

\paragraph{\textbf{Metrics.}}  We report the average precision (AP) with the IoU thresholds at 0.5 and 0.7 on BEV. Each category is divided into easy, moderate, and hard categories according to the height of the 2D bounding box and the occlusion/truncation level.
Specifically, we focus on the ``car" category, following ~\cite{Pseudo-Lidar,ma2019accurate}.

\paragraph{\textbf{Backbone and Local Branch.}}
As used in PointPillars~\cite{pointpillars}, we restrict the point clouds to the detection range of $[-40, 40]m\times[0, 70.4]m\times[-1, 3]m$. The pillar grid size is set to $0.16^2 m^2$.
Since the pseudo-LiDAR points are projected from pixels in the predicted depth map, the maximum number of points per pillar is set to 128, which is 4 times that of the LiDAR points as the input. At the same time, we down-sample the pseudo-LiDAR points 6 times on average.

We use the same backbone and detection head (referred to as the local head in Figure~\ref{fig:pipeline}) in PointPillars.
We add a self-attention module, which is named as $SA_{share}$, on the 4 $\times$ down-sampled pseudo-image of size $(496\times432\times64)$ after extracting the point-level features using PointNet~\cite{pointnet}. Then we concatenate the output of $SA_{share}$ with the features extracted from the 2D backbone in PointPillars. The concatenated feature map is then fed into the two branches for prediction. 

\begin{figure}[t]
\centering
\begin{tabular}{c}
\includegraphics[width=0.8\columnwidth]{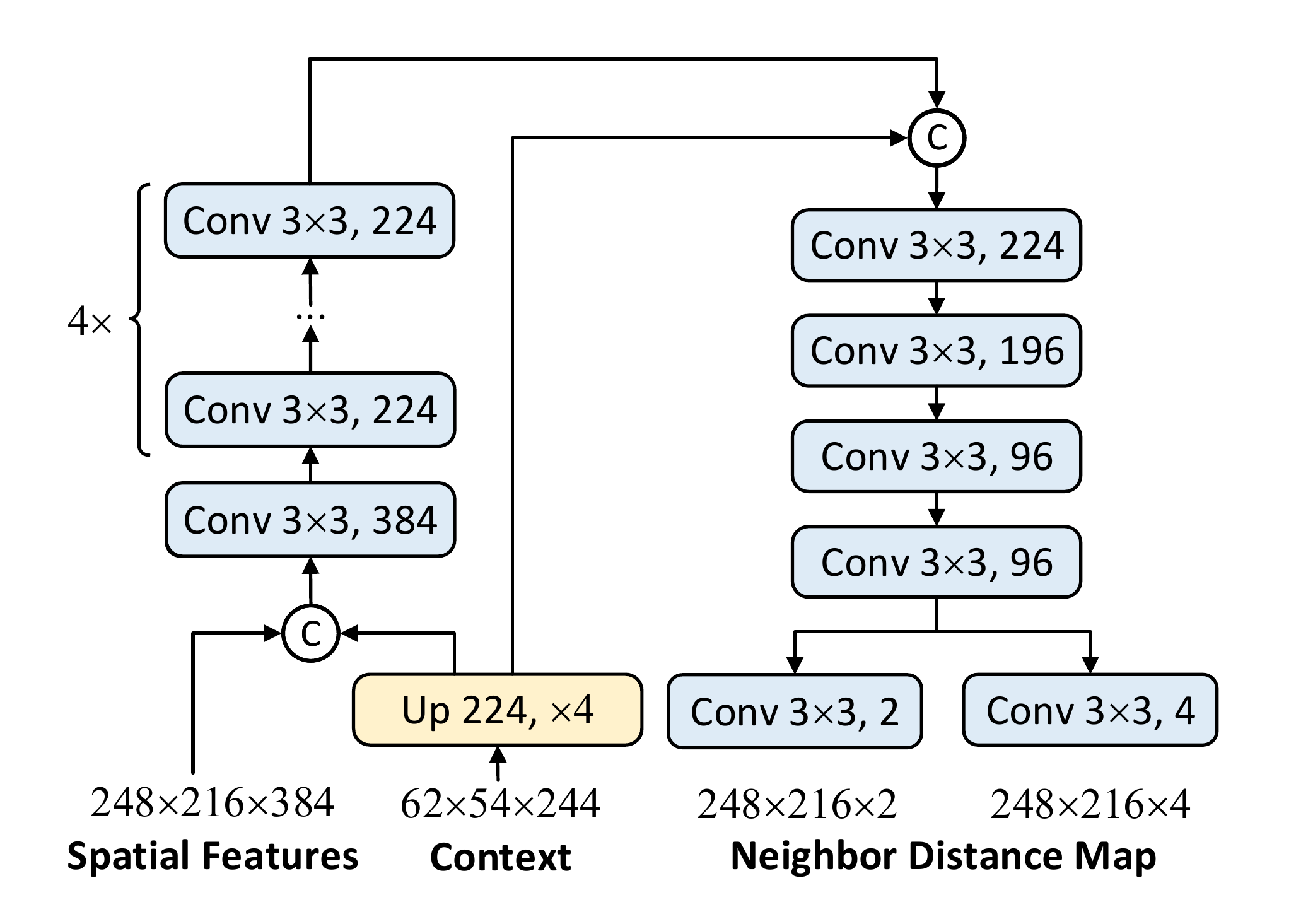}\\
(a) vote head \\
\includegraphics[width=0.7\columnwidth]{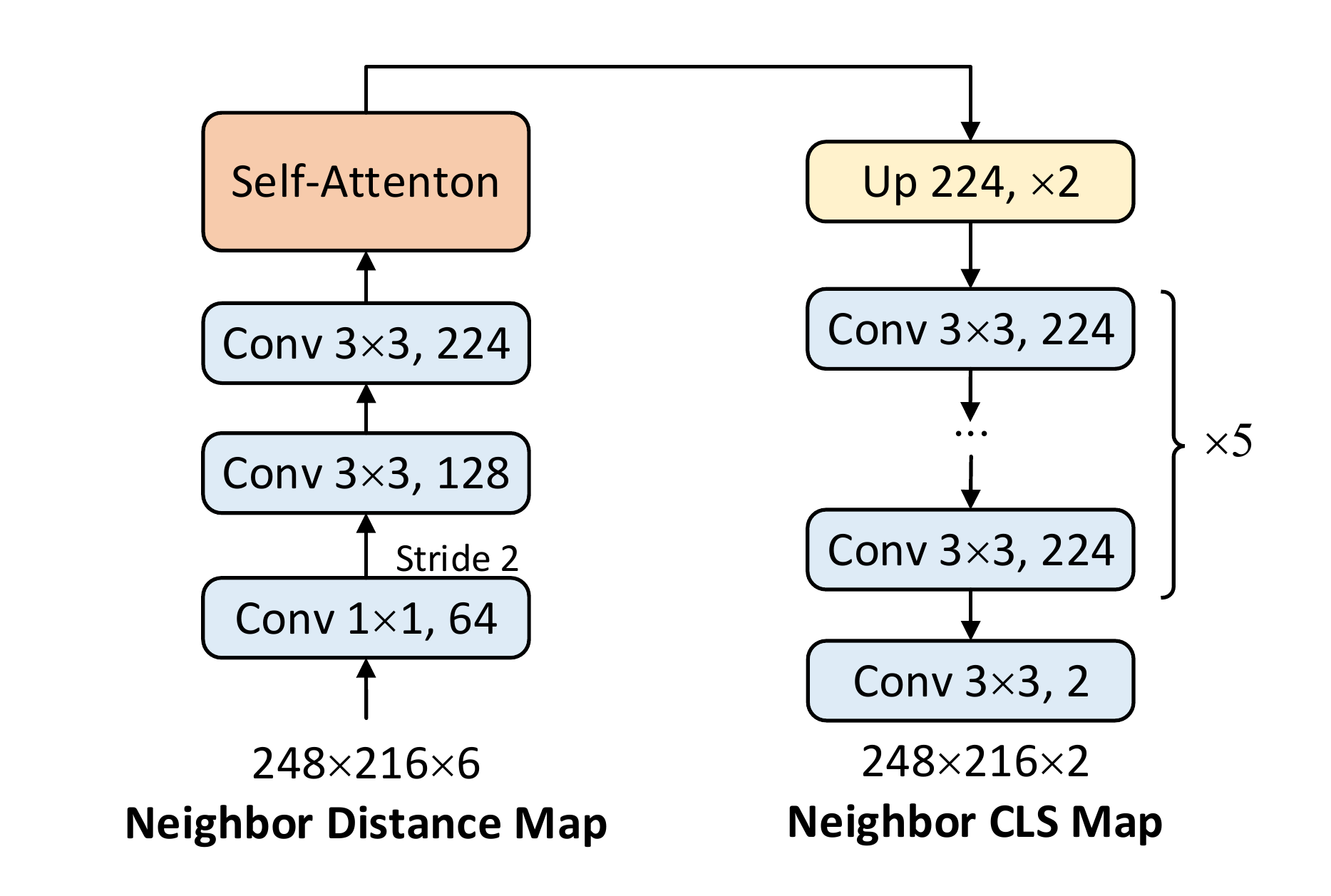} \\ 
(b) neighbor-vote module \\

\end{tabular}
\caption{\label{fig:module} The architecture of our vote head (a) and neighbor-vote module (b). ``Up" means up-sampling, which consists of transposed convolutional layers.  }  
\end{figure}
\paragraph{\textbf{2D ROI Score Association.}}
We apply different 2D detectors to demonstrate the flexibility of ROI score association module. We used FCOS~\cite{FCOS}, CenterNet~\cite{Centernet} and Cascade R-CNN~\cite{cascade}. While these detectors are not the best performers on the COCO~\cite{coco} or KITTI leaderboard, they are open-source and play an important role in improving accuracy in our system. We use the pre-trained weights on COCO dataset and fine-tune it on the 3,712 training images of KITTI. We compare the effect of different 2D detectors on prediction accuracy in Table~\ref{tab:detector}.

\paragraph{\textbf{Neighbor Voting Branch.}}
The structure of our vote head and neighbor-vote module are shown in Figure~\ref{fig:module}. 
In Figure~\ref{fig:module}(a), the spatial features has the dimension of $248\times216\times384$, which is extracted by the 2D backbone.
The context is from the self-attention module $SA_{share}$ in the 2D backbone network. We use $4\times$ down-sampling for better efficiency when applying $SA_{share}$. We predict two tensors and concatenate them together as neighbor distance map. The first 4-channel tensor represents the angle between the object and the positive $X$-axis from the perspective of feature points ( namely ``voters"), denoted by $(sin(\theta_f), cos(\theta_f), sin(\theta_b), cos(\theta_b))$; and the second 2-channel one represents the distance  between the feature points and the object on the $Z$-axis, denoted by $(dz_f, dz_b)$. For supervision, we only take points within $15m$ from the object as valid voters to generate the neighbor distance map as ground-truth. In Figure~\ref{fig:module}(b), we take the neighbor distance map as input, extract features through multiple convolutional layers and a self-attention module, and finally output the neighbor classification map.

\paragraph{\textbf{Monocular Depth Estimation.}} 
Our network is able to adopt any monocular depth estimator. In order to make a fair comparison with other works such as those in~\cite{Pseudo-Lidar,ma2020rethinking,weng2019monocular,ding2020learning} that use depth information, we adopt DORN~\cite{fu2018deep} to predict the depth map. The depth model is trained on about 23,488 images from 32 scenes as in~\cite{eigen2014depth}. Then we use pre-trained weights fixed at training time for depth estimation. 
Since DORN discretizes depth into dozens of categories, the resulting pseudo-LiDAR point clouds present a slice-like distribution.

\begin{figure*}[t]

    \centering
    \includegraphics[width=.98\textwidth]{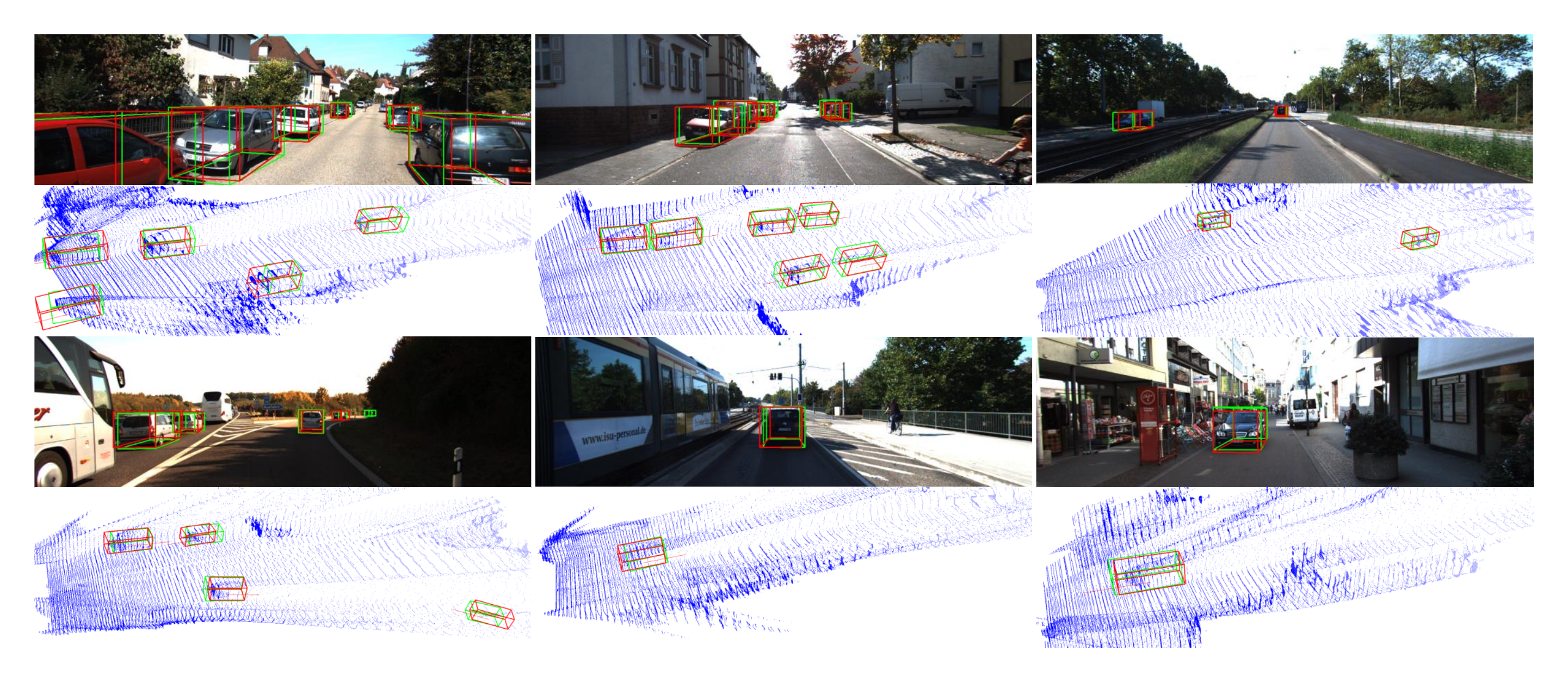}

    \vspace{-6pt}\caption{Qualitative results of our proposed method on KITTI \emph{val} set. We visualize our 3D bounding box estimate (in \textcolor{red}{red}) and ground truth (in \textcolor{green}{green}) on the frontal images (1st and 3rd rows) and pseudo-LiDAR point clouds (2nd and 4th rows). }
    \label{fig:vis}
\end{figure*}
\vspace{-0.35cm}
\begin{table}[t]
	\centering
	
	\small
    \caption{Performance Comparison on KITTI \emph{val} set in terms of $AP_{BEV}$ and $AP_{3D}$ using different 2D detectors in ROI score association module. }
	\begin{center}
            \setlength{\tabcolsep}{4.6pt}
			\begin{tabular}{l|ccc|ccc}
				\toprule
				\multirow{2}{*}{Detector} &  \multicolumn{3}{c}{$\text{Car-AP}_\text{BEV}$ (in \%)~~} & \multicolumn{3}{|c}{$\text{Car-AP}_\text{3D}$ (in \%)~~}\\
				&Easy & Mod. & Hard & Easy& Mod. & Hard\\
				\hline
				CenterNet & 47.1 & 32.8& 27.6& 32.6& 22.7 & 19.8  \\
				Cascade R-CNN& 48.0& 33.0& 28.0 & 32.4& 22.8 & 20.2\\			
				FCOS & 48.2& 33.4 & 28.2 & 33.4 & 23.4 & 20.1 \\

			\bottomrule
			\end{tabular}
	\end{center}
	\label{tab:detector}
\end{table}

\paragraph{\textbf{Training Stage.}} We focus on the detection accuracy on the bird's eye view. The training process is divided into two steps. We first remove the neighbor-voting branch, but still predict the neighbor distance map and calculate the loss to back propagate the gradients. We train this network for 65 epochs and then use the pre-trained weights to train the whole network (including the neighbor-voting branch) for 60 epochs. During training, for the detection loss, the hyper-parameters follow the same settings as used in Voxel R-CNN~\cite{voxel-rcnn}, PV-RCNN~\cite{pvrcnn} and OpenPCDet~\cite{openpcdet2020}. For the voting loss, we empirically set the hyper-parameters for $L\_{dist}$ and $L\_{ang}$ as 0.2 and 0.06 to keep the amplitude of each loss comparable.
The whole architecture of our network is end-to-end optimized with the Adam optimizer with the batch size 12. The learning rate is initialized as 0.03 for the first step, 0.02 for the second step and both updated by cosine annealing strategy. For data augmentation, we apply random mirroring flip along the $Z$ axis and rotation between $[-5, 5]$ degrees along the $Y$ axis.

For comparison, we also train a \emph{baseline} network, which consists only of a pseudo-LiDAR generation module and a LiDAR-based 3D detector same as PointPillars~\cite{pointpillars}.

\paragraph{\textbf{Inference Stage.}} At the inference stage, we input the original neighbor distance map into the neighbor-vote module without restricting the distance to the voted objects. Non maximum suppression (NMS) is taken with IoU threshold 0.25 to remove the redundant predictions. 

\subsection{Evaluation Results}
We report our performance on both \emph{val} set and \emph{test} set for comparison and analysis. The performance on \emph{val} set is calculated with the AP setting of recall 11 positions to compare with the previous works. And the results evaluated by the test server utilize AP setting of recall 40 positions.

\paragraph{\textbf{Comparison Results on the Validation Set.}}
We first compare the BEV and 3D detection accuracy of \systemname\ with several recent monocular networks that represent the state of the art on the validation dataset. These networks fall into two categories: (1) networks that require extra information such as depth or instance segmentation, and (2) networks that do not require any extra information. \systemname\ belongs to the former category, and the networks in this category usually perform better. The results are summarized in Table~\ref{table:eval_bev}. The results show that \systemname\ performs the best among all the methods. For example, compared to the current best monocular network on the validation set, \emph{i.e.}, PatchNet~\cite{ma2020rethinking}, we improve the absolute $AP_{BEV}$ in the easy, moderate, and hard category by $3.8\%$, $4.3\%$ and $4.1\%$ with $IoU=0.7$, with the corresponding relative improvement of $8.6\%$, $14.8\%$ and $17.0\%$. Examples of our 3D bounding box estimate on KITTI \emph{val} set are visualized in Figure~\ref{fig:vis}.
\begin{table}[tb]
\vspace{-0.25cm}
	\centering
	\small
    \caption{Performance Comparison on KITTI \emph{test} set in terms of $AP_{BEV}$. ``Extra Information'' means extra supervision in addition to the ground-truth boxes, where ``mask" means the label of segmentation task.}
	\begin{center}
			\setlength{\tabcolsep}{4.5pt}
			\begin{tabular}{l|cc|rrr}
				\toprule
				\multirow{2}{*}{Method} & \multicolumn{2}{c}{ Extra Info} &  \multicolumn{3}{|c}{$\text{Car-AP}_\text{BEV}$ (in \%)~~} \\ 
				&Depth&Mask&Easy & Mod. & Hard \\
				\hline
				$\text{ROI-10D}$~\cite{manhardt2019roi} &\checkmark& & 9.78 & 4.91& 3.74 \\
				$\text{OAV}$~\cite{OAV} & & & 16.24 & 10.13& 8.28 \\				
				$\text{MonoGRNet}$~\cite{qin2019monogrnet}&\checkmark& & 18.19 & 11.17& 8.73 \\				
				$\text{Mono3D-PL}$~\cite{weng2019monocular}& \checkmark& \checkmark& 21.27  & 13.92  & 11.25 \\
				$\text{RTM3D}$~\cite{li2020rtm3d}& & & 19.17  & 14.20  & 11.99 \\
				$\text{SMOKE}$~\cite{liu2020smoke}& & & 20.83 & 14.49 & 12.75\\	
				$\text{D4LCN}$~\cite{ding2020learning}&\checkmark& & 22.51 & 16.02 &12.55 \\
				
				$\text{PatchNet}$~\cite{ma2020rethinking}&\checkmark& & 22.97 & 16.86 & 14.97\\
				$\text{AM3D}$~\cite{ma2019accurate}& \checkmark & \checkmark&25.03 & 17.32 & 14.91\\
                \hline
                \hline
                \textbf{\systemname$\text{(ours)}$}&\checkmark&
                &\textbf{27.39}  & \textbf{18.65} & \textbf{16.54}  \\
                
			\bottomrule
			\end{tabular}
	\end{center}
	\label{tab:test}
\vspace{-0.25cm}
\end{table}

\begin{table}[t]
\centering
\small
\caption{Ablation analysis on KITTI \emph{val} set. We quantify individual and combined impacts of $\text{SA}_\text{share}$ (SA), ROI score association (RA), neighbor-voting branch (V), and the fusion of the two classification branches (F).} 
\setlength{\tabcolsep}{4.6pt}
\begin{tabular}{cccc|rrr|rrr}
\toprule
    \multirow{2}{*}{SA} & \multirow{2}{*}{RA} & \multirow{2}{*}{V} &\multirow{2}{*}{F} & \multicolumn{3}{c|}{$\text{AP}_{\text{BEV}}$  (\%)} & \multicolumn{3}{c}{$\text{AP}_{\text{3D}}$  (\%)}\\
    & & & & Easy & Mod. & Hard & Easy & Mod. & Hard \\
    \hline 

     & & & & 39.1 & 26.9 & 24.6 & 25.3 & 17.2 & 15.9\\ 

    \checkmark & & & &  41.8 & 28.2 & 25.5 & 30.1& 20.3 & 18.8 \\ 
    \checkmark &  & \checkmark& & 46.2 & 30.6 & 27.1 & 30.8 & 21.8 & 18.2 \\ 
    \checkmark & &\checkmark & \checkmark & 46.9 & 31.2 & 27.4 & 31.3 & 21.7  & 18.6 \\  
    \checkmark&\checkmark&\checkmark&\checkmark & \textbf{48.2}& \textbf{33.4} & \textbf{28.2} & \textbf{33.4} & \textbf{23.4} & \textbf{20.1}\\

\bottomrule
\end{tabular}
\vspace{-0.25cm}
\label{tab:ablation}
\end{table}
\paragraph{\textbf{Comparison Results on the Test Set.}}
We next show the \emph{test} results in Table~\ref{tab:test}. Among all the monocular networks, \systemname\ is the best, with $9.4\%$, $7.7\%$ and $10.9\%$ relative improvement in the easy, moderate, and hard category compared to the current best monocular network on the \emph{test} set AM3D~\cite{ma2019accurate}. 

\paragraph{\textbf{Effectiveness of Neighbor Voting.}}
Table~\ref{tab:ablation} also shows the effectiveness of the neighbor-voting branch. For example, it (in the third row) can improve the $AP_{BEV}$ by 4.4\%, 2.4\%, 1.6\% for easy, moderate and hard level detection with $IoU=0.7$, compared with the version without neighbor-voting (in the second row). 
It is noted that at this time, we only add the neighbor-voting branch and use the supervision of the neighbor distance map without fusion with the local branch. That is, only the classification score of the local branch is used. Even so, our network has a significant improvement on $AP_{BEV}$. After fusion of two branches, the network achieves a slightly better performance.

This shows that neighbor-voting can effectively improve the detection accuracy from deformed pseudo-LiDAR point clouds. Specifically, as discussed earlier, the deformation in the pseudo-LiDAR point clouds becomes worse as the distance increases. Fortunately, as shown in Table~\ref{tab:distance}, the relative performance improvement from neighbor-voting is the greatest with long distances, \emph{i.e.}, 253\% on $AP_{BEV}$ with $\sim70m$. This again confirms that neighbor-voting is very effective in improving the detection accuracy, even when the pseudo-LiDAR point clouds are seriously deformed.  

\begin{table}[t]
\centering
\small
\caption{Performance Comparison on KITTI \emph{val} set in terms of $AP_{BEV}$ and $AP_{3D}$ for different object distance ranges with and without neighbor-voting. The relative improvement (\%) is also included.}
\setlength{\tabcolsep}{4.6pt}
\begin{tabular}{l|cc|cc}
\toprule
    \multirow{2}{*}{Method} & \multicolumn{2}{c}{Hard $\text{AP}_{\text{BEV}}$ (in \%)} & \multicolumn{2}{|c}{Hard $\text{AP}_{\text{3D}}$ (in \%)} \\
    & $0 \sim40m$ & $40 \sim 70m$& $0 \sim40m$ & $40 \sim 70m$ \\
    \hline
    no voting & 26.48 & 1.29 & 16.65& 0.76 \\
    with voting & 29.47& 4.55& 20.88 & 3.03 \\
    \hline
    \hline
    Impro. (\%) & +11.3 & +253  & +25.4& +299 \\
\bottomrule
\end{tabular}

\label{tab:distance}
\end{table}

\begin{table}[t]

\centering
\small
\caption{Relative changes in the number of false positive
boxes and the number of true positive boxes on KITTI \emph{val} set.}
\setlength{\tabcolsep}{5pt}
\begin{tabular}{l|cc|cc|cc}
\toprule
    \multirow{2}{*}{Method} &  \multicolumn{2}{c|}{$IoU^*=0.3$} &  \multicolumn{2}{c|}{$IoU^*=0.5$} &  \multicolumn{2}{c}{$IoU^*=0.7$} \\
    &$N_{tp}$ & $N_{fp}$&$N_{tp}$ & $N_{fp}$ &$N_{tp}$ & $N_{fp}$\\
    \hline
    Baseline &8395 &2046 & 8190&2364 & 7181& 3646   \\
    Ours & 8457& 859 & 8306 &1106 &7356 &2324 \\
    \hline
    \hline
    \tabincell{l}{Changes (\%)} & +0.7 & -58   & +1.4 & -53 &  +2.4  & -36   \\
\bottomrule
\end{tabular}

\label{tab:num_fp}
\end{table}
\paragraph{\textbf{Effectiveness of \systemname in Reducing the False-Positive Rate.}}
The underlying rationale behind \systemname is that we believe most feature points would vote for true objects. As such, the neighbor-voting mechanism can effectively filter out  false-positive predictions. In order to confirm this rationale, we compare the number of true-positives and false-positives with different IoU thresholds in the \emph{baseline} network and our network, as shown in Table~\ref{tab:num_fp}. Specifically, when the IoU between a predicted box and the ground truth is greater than a preset threshold, \emph{e.g.} 0.3, 0.5 or 0.7, this predicted box is considered a true positive (TP); otherwise it is a false positive (FP).  

\begin{figure}[t]
\centering
\begin{tabular}{cc}
\includegraphics[width=0.48\columnwidth]{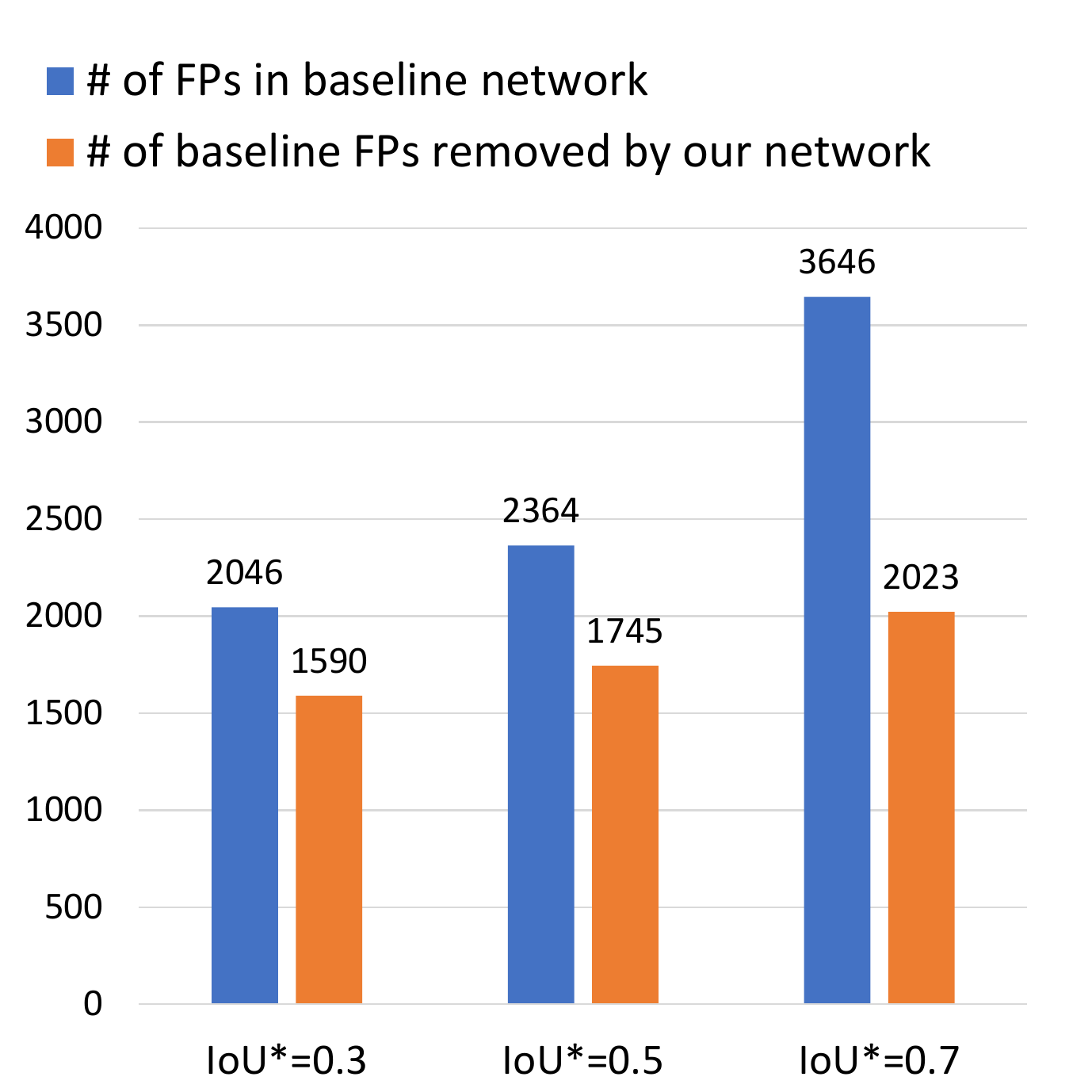}&
\includegraphics[width=0.48\columnwidth]{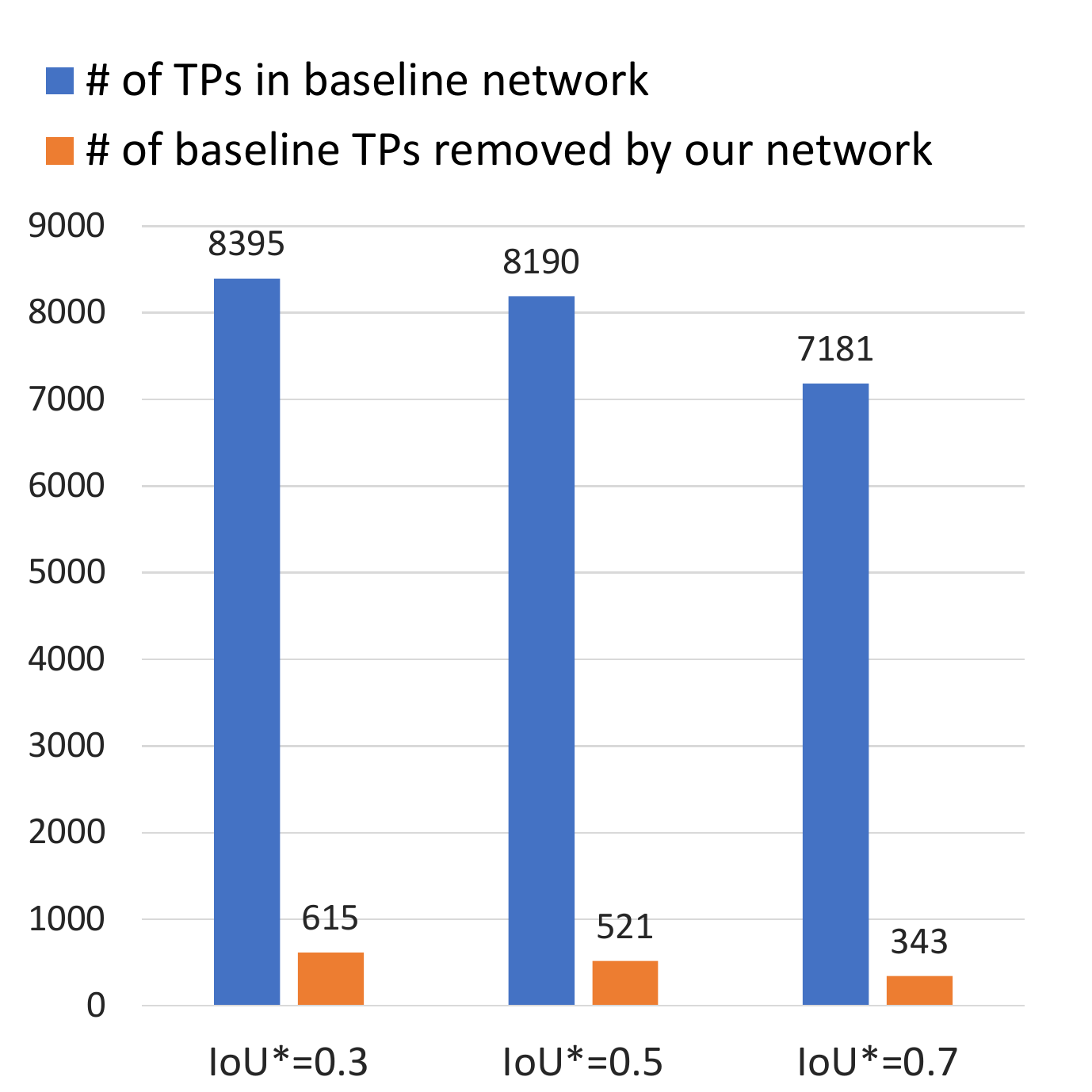} \\ 
(a) FP & (b) TP\\
\end{tabular}
\caption{\label{fig:fp_tp_remove} We report the number of FPs in the \emph{baseline} network, and the number of baseline FPs successfully removed in our network in (a);  the number of TPs in the \emph{baseline} network, and the number of baseline TPs accidentally removed in our network in (b).} 
\vspace{-0.2cm}
\end{figure}

Next, we count the number of FP boxes that are in the \emph{baseline} network but not in our network. Here, we set the IoU threshold to be 0.1 -- when the IoU of two boxes is greater than 0.1, the two boxes are considered the same. In this way, we are reporting the \emph{lower bound} of the FP boxes that are effectively removed by our network. The results show that our network eliminates 73.8\% and 55.4\% of the FP boxes with $IoU^*=0.5$ and $IoU^*=0.7$ on the KITTI \emph{val} set. Finally, our network does introduce new FP boxes, but the number is rather small, \emph{i.e.}, the new FP boxes only account for 25\% of the eliminated FP boxes with $IoU^*=0.3$. The more detailed results are shown in Figure~\ref{fig:fp_tp_remove}(a). 

Finally, we also investigate whether \systemname also removes TP boxes that are generated by the \emph{baseline} network. As shown in Figure~\ref{fig:fp_tp_remove}(b), only a small portion of TPs are lost in our network, \emph{e.g.} 6.4\% and 4.8\% with $IoU^*=0.5$ and $IoU^*=0.7$ on the KITTI \emph{val} set.
\vspace{-0.15cm}
\paragraph{\textbf{Effectiveness of Self-Attention.}}
Besides using self-attention in the neighbor-vote module, we also use it in the 2D backbone, which is named as $SA_{share}$. According to Table~\ref{tab:ablation}, $SA_{share}$ contributes to the overall detection accuracy. This confirms the effectiveness of spatial context retrieval in the feature extraction step.
\vspace{-0.15cm}
\paragraph{\textbf{Effectiveness of 2D ROI Score Association.}}
We use FCOS as the 2D detector for ROI score association.
Comparing the last two lines in Table~\ref{tab:ablation}, the ROI score association improves $AP_{3D}$ by 1.7\% for the moderate category and by 2.1\% for the easy category.

\section{Conclusion}
In this work, we present \systemname\ for monocular 3D object detection.  The key difference from prior works lies in the fact that we take into consideration the predictions of the neighbor feature points around the objects to help ameliorate detection from severely deformed pseudo-LiDAR. The individual, noisy predictions of each feature point in the neighborhood can collectively form a much informed prediction through voting. Further, final predictions are obtained by combining the neighborhood predictions with local predictions via adaptive weights. Experiments on KITTI Dataset prove the effectiveness of the proposed method, which outperforms the state-of-the-art monocular networks by a noticeable margin.

{\small
\bibliographystyle{ieee_fullname}
\bibliography{egbib}
}

\clearpage

\end{document}